%% file: main.tex
\documentclass[runningheads]{llncs}

\usepackage{eccv}

\usepackage{eccvabbrv}

\usepackage{graphicx}
\usepackage{booktabs}

\usepackage[accsupp]{axessibility}  

\usepackage{hyperref}

\usepackage{orcidlink}

\usepackage{xspace}
\usepackage{enumitem}
\usepackage{booktabs}
\usepackage{multirow}
\usepackage{graphicx}
\usepackage[utf8]{inputenc}
\usepackage{xcolor}
\usepackage{subcaption}
\usepackage{wrapfig}
\usepackage{colortbl}

\definecolor{lightgreen}{RGB}{230, 255, 230}
\definecolor{lightred}{RGB}{255, 230, 230}
\definecolor{maxblue}{RGB}{200, 230, 255} 
\definecolor{minred}{RGB}{255, 200, 200}  

\newcommand{\bench}{DiCoBench\xspace}

\begin{document}

\title{DiCoBench: Benchmarking Multi-Image Fine-Grained Perception via Differential and Commonality Visual Cues} 

\titlerunning{DiCoBench}




\author{Geng Li\inst{1}\orcidlink{0000-0002-9733-6891} \and
Yuxin Peng\inst{1}\thanks{Corresponding author.}\orcidlink{0000-0001-7658-3845}}

\authorrunning{G.~Li and Y.~Peng}

\institute{Wangxuan Institute of Computer Technology, Peking University, Beijing, China\\
\email{ligeng@stu.pku.edu.cn, pengyuxin@pku.edu.cn}}

\maketitle

\begin{abstract}
  Recent advancements in Multimodal Large Language Models (MLLMs) have demonstrated impressive fine-grained perception capabilities. However, existing benchmarks predominantly rely on explicit textual cues or low-resolution inputs, failing to evaluate a model's ability to autonomously perceive implicit visual cues in high-resolution. To bridge this gap, we introduce \textbf{DiCoBench}, a comprehensive, multi-image high-resolution benchmark designed for cross-image fine-grained perception. DiCoBench consists of 765 meticulously curated samples categorized into two progressive tracks: Differential Visual Cues and Commonality Visual Cues, covering 8 distinct perception tasks. By formulating the benchmark as a multiple-choice question task and utilizing high-resolution imagery (approaching 2K), we eliminate evaluation metric bias and pose a substantial challenge to current state-of-the-art MLLMs. Our extensive evaluation of 18 diverse MLLMs reveals a striking performance gap compared to human accuracy (98.3\%), with top-performing models struggling significantly with  micro-scale detail capture. We believe DiCoBench will serve as a challenging testbed to drive future research in autonomous, high-resolution multi-image perception.
\end{abstract}

\input{sections/1_introduction}
\input{sections/2_related_works}
\input{sections/3_bench}

\input{sections/5_experiments}

\input{sections/6_conclusion}


\section*{Acknowledgements}
This work was supported by the grants from the National Natural Science Foundation of China (62525201, 62132001, 62432001) and Beijing Natural Science Foundation (L247006, L257005).

%
%
\bibliographystyle{splncs04}
\bibliography{main}
\end{document}

%% file: sections/1_introduction.tex
\section{Introduction}
\label{sec:intro}

\begin{figure}[htbp]
    \centering
    \newcommand{\figheight}{3.8cm} 

    \begin{subfigure}[b]{0.32\textwidth}
        \centering
        \includegraphics[height=\figheight, width=\textwidth]{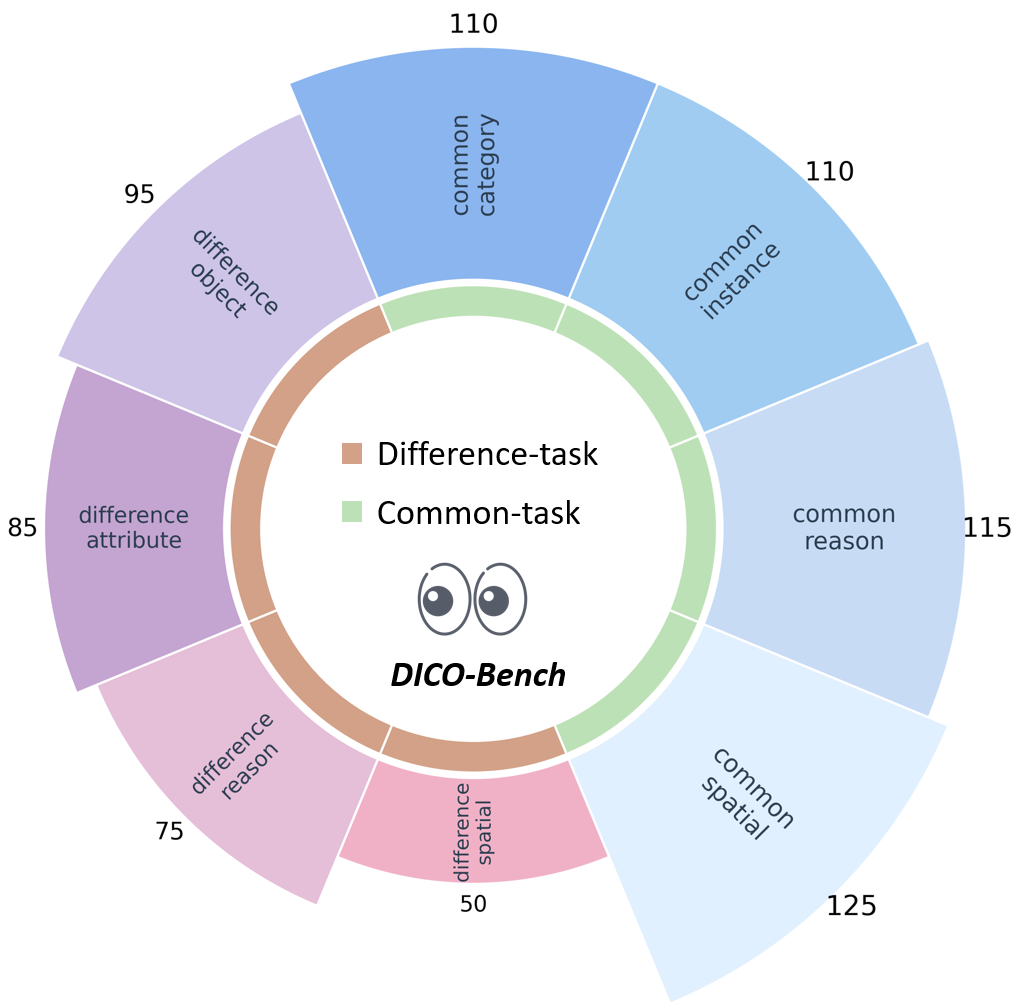}
        \caption{Composition of our proposed \bench.}
        \label{fig:right}
    \end{subfigure}
    \hfill 
    \begin{subfigure}[b]{0.32\textwidth}
        \centering
        \includegraphics[height=\figheight, width=\textwidth]{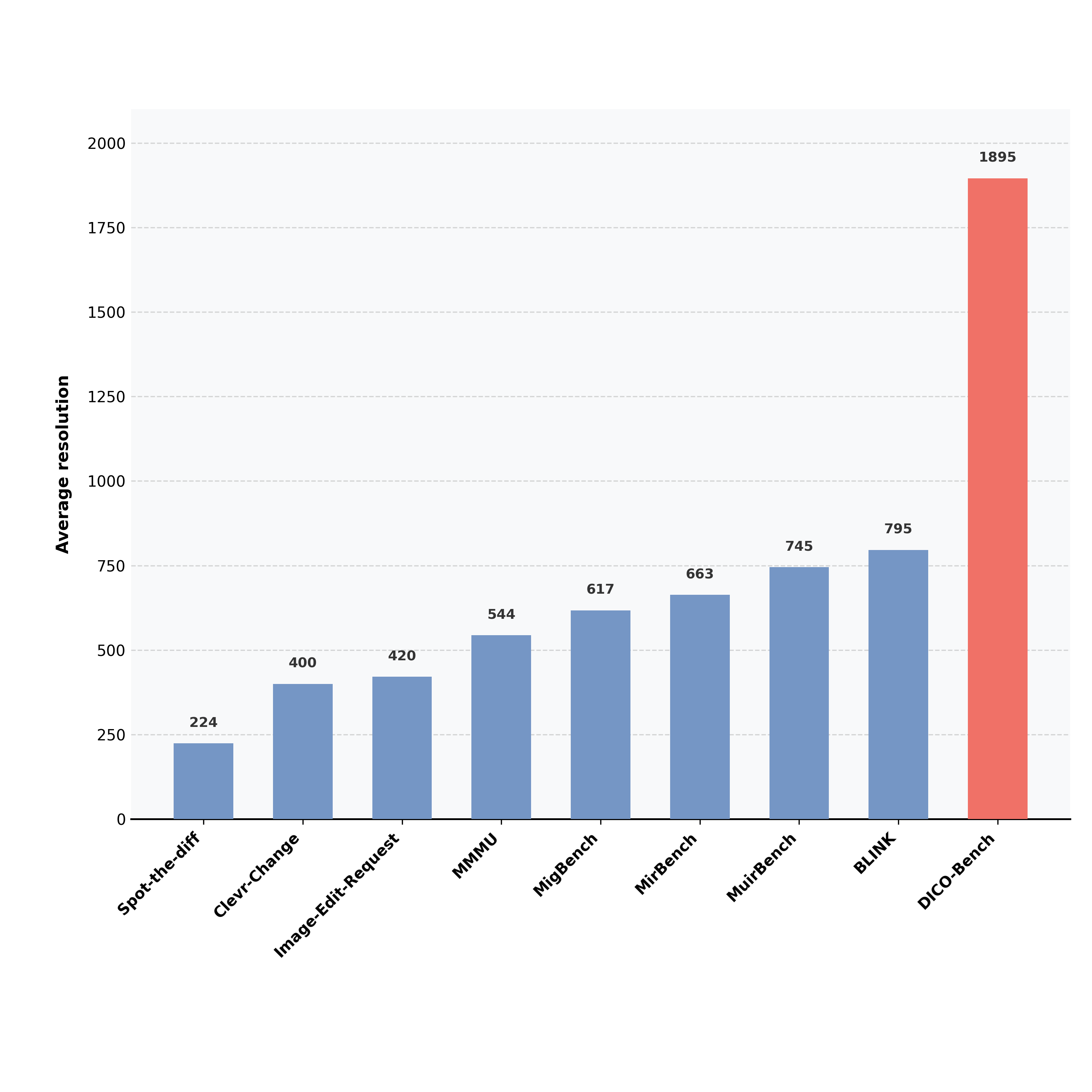}
        \caption{Comparison of image resolutions across multi-image multimodal benchmarks.}
        \label{fig:left}
    \end{subfigure}
    \hfill 
    \begin{subfigure}[b]{0.32\textwidth}
        \centering
        \includegraphics[height=\figheight, width=\textwidth]{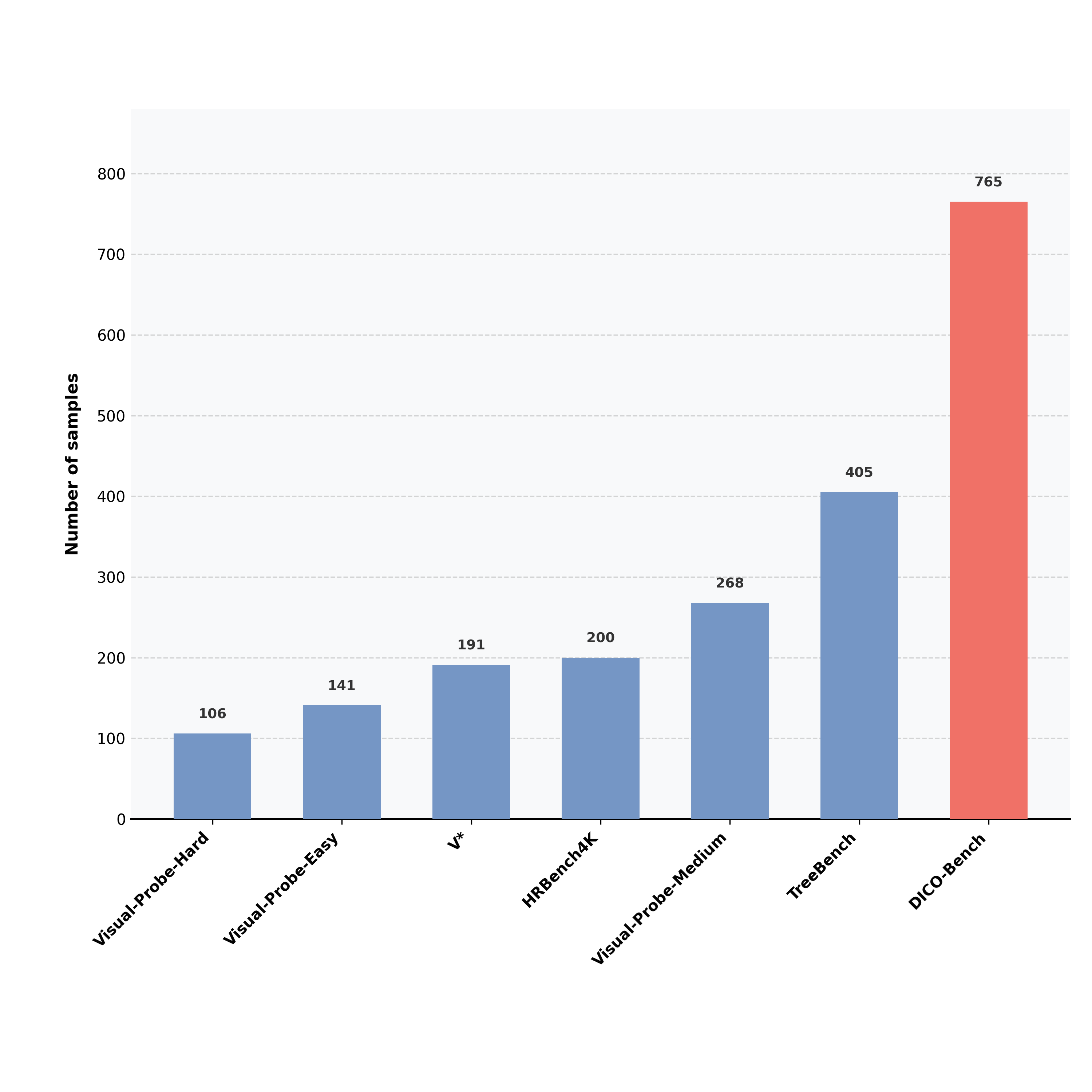}
        \caption{Comparison of sample sizes across fine-grained perception benchmarks.}
        \label{fig:middle}
    \end{subfigure}
    
    \caption{\textbf{Overview of our proposed \bench}. (a) \bench covers 2 major perception categories and 8 specific perception tasks. (b) We observe that the average resolution of existing multi-modal benchmarks remains primarily in low-resolution scenarios. In contrast, our proposed \bench reaches approaching 2K. (c) Due to high-resolution constraints, the largest existing single-image fine-grained perception benchmarks contain only around 400 samples. \bench with a sample size and diversity nearly double that of the current largest benchmark. }
    \label{fig:first}
\end{figure}

\begin{figure}[tb]
  \centering
  \includegraphics[width=\textwidth]{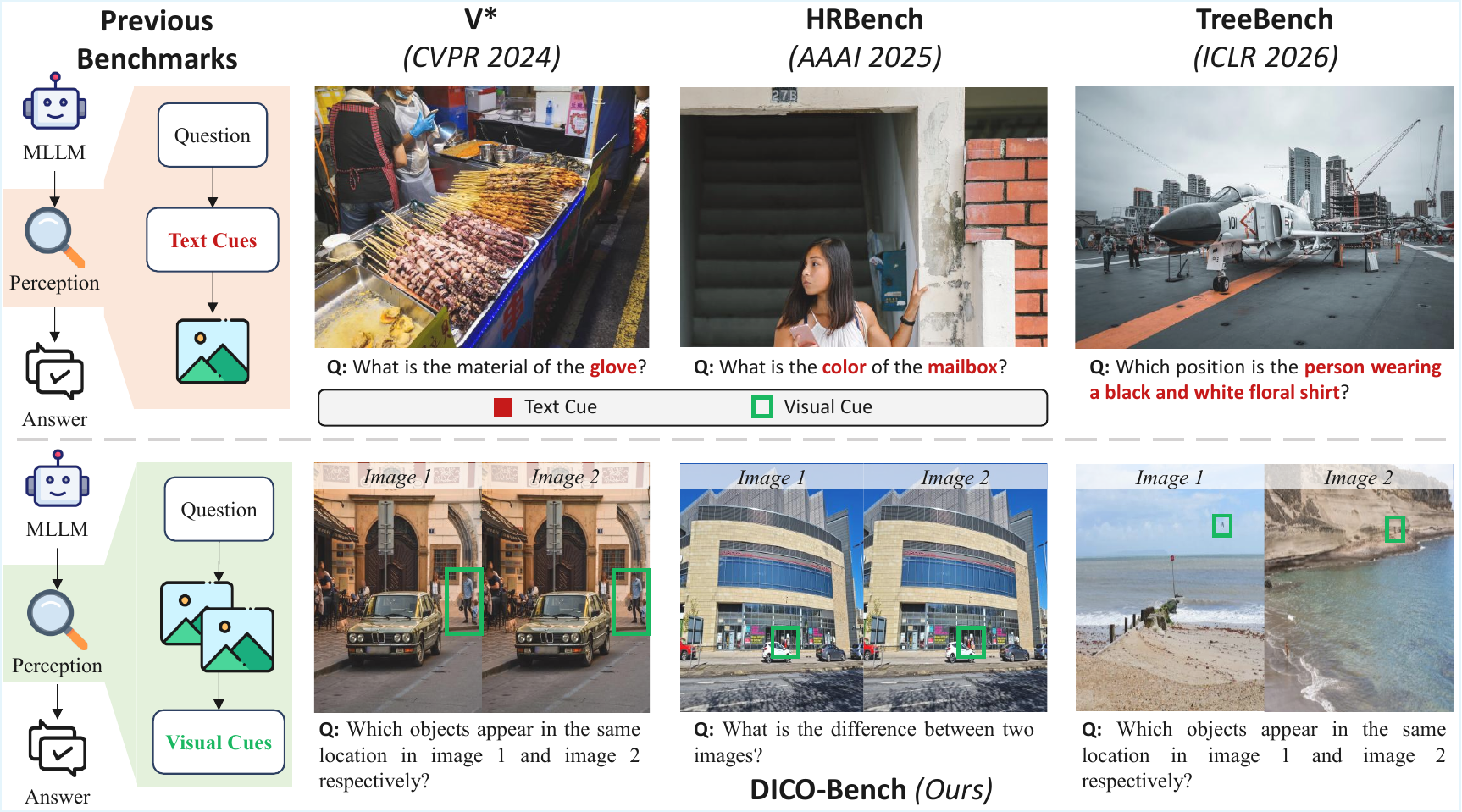}
  \caption{\textbf{Comparison between \bench and previous benchmarks.} Previous fine-grained perception benchmarks primarily rely on explicit textual prompts (highlighted in red) related to objects to guide perception. In contrast, the key distinction of \bench is its emphasis on evaluating a model's ability to drive multi-image perception through implicit visual cues of commonalities and differences (highlighted in green boxes), which more closely resembles the human perceptual process in dynamic, real-world environments without text cues available.}
  \label{fig:bench_compare}
\end{figure}

Driven by the evolution of high-resolution visual encoders and massive vision-language alignment data, recent Multimodal Large Language Models (MLLMs) have achieved remarkable breakthroughs in single-image fine-grained perception~\cite{bai2025qwen25vl,bai2025qwen3vl,an2025llavaonevision15,zheng2025deepeyes,hong2026deepeyesv2,zhang2025thyme,wang2025treebench,lai2025minio3}. 
As demonstrated by recent benchmarks like V*~\cite{wu2024vstar}, HR-Bench~\cite{wang2025hrbench}, TreeBench~\cite{wang2025treebench} and Visual Probe~\cite{lai2025minio3}, advanced MLLMs exhibit remarkable proficiency in localizing and recognizing minute details in single high-resolution image. 
However, these existing fine-grained evaluations fundamentally rely on explicit textual cues contained by question (\textit{e.g.}, ``Where is the \textit{umbrella}?'' or ``Who the \textit{person in red} is?''). 
Essentially, they mainly assess a model's passive text-to-image grounding capability. 
In contrast, real-world visual cognition is rarely instruction-driven; when navigating complex, dynamic environments, human perception proactively captures spontaneous \textit{implicit visual cues}, such as subtle visual differences or commonalities across multiple observations. 
Current single-image benchmarks fail to measure this ``autonomous visual-cue guided perception.'' Consequently, we find when deployed without explicit textual guidance and forced to discover extremely minute visual cues purely through cross-image comparison, even state-of-the-art MLLMs suffer from severe performance degradation.

To address the limitations of single-image evaluation, the community has increasingly focused on multi-image multimodal benchmarks. 
Comprehensive benchmarks like MuirBench~\cite{wang2024muirbench}, MIR Benchmark~\cite{DuEtAl2025mirbench}, BLINK~\cite{fu2024blink}, and MLLM-COMPBENCH~\cite{kil2024mllmcompbench} evaluate broad capabilities such as  STEM knowledge, scene understanding and temporal ordering. 
Concurrently, domains like Image Difference Captioning (IDC)~\cite{jhamtani2018spot_the_diff, tan2019ier, park2019clevr_change, di2025difftell, liu2025omnidiff} require models to describe local changes between image pairs. 
Despite these advancements, existing multi-image multimodal benchmarks face three bottlenecks. 

1. \textit{Evaluation Metric Bias}: IDC tasks predominantly frame perception as text generation, evaluated by strict n-gram matching metrics like ROUGE-L or CIDEr. As demonstrated by G-VEval~\cite{tong2025gveval}, these metrics severely penalize MLLMs for formatting differences rather than genuine perception errors, thereby suppressing and misrepresenting the true perceptual limits of advanced models like Qwen2-VL~\cite{wang2024qwen2vl}. 

2. \textit{Lack of Systematic Taxonomy}: Existing multi-image benchmarks fail to consider implicit visual cues as a systematic evaluation dimension, resulting in fragmented and unsystematic assessments. For instance, tasks like IDC exclusively focus on capturing visual differences, completely neglecting the equally crucial cognitive dimension of visual commonalities.

3. \textit{Insufficient Image Resolution}: Current multi-image multimodal datasets predominantly rely on low-resolution scenes like~\Cref{fig:first}. They severely lack the capacity to evaluate a model's perceptual limits under high-resolution conditions, making it extremely difficult to ascertain whether models can genuinely capture minute visual cues to achieve accurate fine-grained perception.

To bridge the gap mentioned above, we propose a novel and highly challenging task: \textit{Vision-cue guided Cross-Image Fine-grained Perception}, accompanied by a new multi-image multimodal benchmark, \bench~(Difference-Common Bench). Innovatively, \bench systematically categorizes the visual cues captured by humans without text guidance into two parallel tracks: \textit{Differential Visual Cues} and \textit{Commonality Visual Cues}. To comprehensively evaluate a model's ability to acquire these cues, we specifically design 4 perception tasks for each track. This yields a total of 8 distinct task categories, ranging from attributes and instances to spatial relationships and logical reasoning.
To fundamentally eliminate the evaluation metric bias caused by n-gram text generation penalties, we formulate \bench as a Multiple-Choice Question (MCQ) task, universally appending a robust ``No visible difference'' or ``No visible commons'' option to complete the logical space. 

To ensure the images are sufficiently high-definition to pose a substantial challenge to existing models, \bench collects base images with average resolution of 1895, approaching 2K.

We systematically evaluate 18 MLLMs of diverse architectures and scales on the \bench. Our results reveal a striking paradox: \textbf{while these tasks are intuitive and trivial for humans (98.3\% average accuracy), they remain exceptionally challenging for current SOTA models}. Even the most capable closed-source model, Gemini-3-Pro, achieves only 58.1\% average accuracy, trailing behind human performance by 40.2\%, and representing only a modest advancement over the capabilities of current open-source alternatives. Furthermore, we observe significant performance volatility across different task types; for instance, while models demonstrate emerging proficiency in categorical tasks, their ability to conduct high-level logical reasoning during perception remains severely constrained, with scores often plummeting toward random-guessing levels. Our findings indicate that the cross-image fine-grained perceptual capabilities of current MLLMs have been significantly overestimated. We believe \bench highlights a critical frontier in MLLM perception, serving as an effective testbed to bridge the profound gap in high-resolution, visual-cue guided, and multi-image perception.

%% file: sections/2_related_works.tex
\section{Related Works}
\label{sec:related_work}

\subsection{Text-Guided Fine-Grained Perception}
Driven by the evolution of high-resolution visual encoders and dynamic spatial pooling strategies (\textit{e.g.}, Qwen2-VL~\cite{wang2024qwen2vl}, InternVL-1.5~\cite{chen2024internvl15}), recent MLLMs have demonstrated remarkable proficiency in perception.
Consequently, benchmarks such as V*~\cite{wu2024vstar}, HR-Bench~\cite{wang2025hrbench}, TreeBench~\cite{wang2025treebench}, and Visual Probe~\cite{lai2025minio3} have been proposed to further evaluate the localization and recognition of minute details in high-resolution images, also known as the fine-grained perception task.
However, these benchmarks fundamentally rely on \textit{explicit textual cues} (\textit{e.g.}, ``Where is the red cup in the image?'').
They evaluate a model's passive text-to-image grounding capability rather than its proactive ability to mine and interpret spontaneous visual cues in the wild.
We find that although existing MLLMs have already performed excellently in the aforementioned tasks (e.g., the best model on V* has reached 90\% accuracy), when deployed in complex, dynamic environments without explicit textual guidance, existing MLLMs often suffer from severe performance degradation, highlighting a critical gap in autonomous high-resolution fine-grained perception.

\subsection{Multi-Image Perception}
%
To push MLLMs beyond single-image constraints, researchers have increasingly focused on multi-image perception and reasoning tasks.
Comprehensive benchmarks such as MMMU~\cite{yue2024mmmu}, BLINK~\cite{fu2024blink}, MuirBench~\cite{wang2024muirbench}, MLLM-CompBench~\cite{kil2024mllmcompbench}, MileBench~\cite{song2024milebench} and MIR Benchmark~\cite{DuEtAl2025mirbench}  have been introduced to evaluate broad capabilities, including STEM knowledge, scene understanding, multi-view reasoning and \etc. 
Concurrently, specialized datasets have emerged to probe specific multi-image perception dimensions: MIG-Bench~\cite{li2025migbench} explores free-form multi-image grounding, while MIHBench~\cite{li2025mihbench} specifically evaluates multi-image hallucinations regarding object existence and identity consistency.
Despite these advancements, existing multi-image perception benchmarks face structural limitations. 
They rely on low-resolution images, which inevitably lose fine-grained visual details. 
More importantly, their task designs predominantly focus on macro-level logical relationships or the association of highly salient objects, inherently failing to assess the fine-grained ability to discover micro-level visual cues under high-resolution, complex background interference.

\subsection{Image Difference Captioning}
%
The most closely related domain to our work is ``spot-the-difference'' or Image Difference Captioning (IDC).
Early foundational works introduced datasets like CLEVR-Change~\cite{park2019clevr_change}, Spot-the-Diff~\cite{jhamtani2018spot_the_diff}, and Image Editing Request (IER)~\cite{tan2019ier} benchmarks to explore pixel-level modifications or synthetic visual changes.
Building upon this, recent efforts have constructed various datasets to train and evaluate MLLMs on capturing realistic visual changes, including DiffTell~\cite{di2025difftell} for image manipulations, M3-Verse~\cite{wei2025m3verse} for 3D environments, OmniDiff~\cite{liu2025omnidiff}, and OneDiff~\cite{hu2024onediff}. 
Other works~\cite{zhang2024differentialperceptive, guo-etal-2025-learning-describe, li-etal-2025-change} have focused on enhancing MLLMs' architectures or pre-training strategies for difference captioning. VisMin~\cite{awal2024vismin} further introduces visual minimal-change understanding.
However, this paradigm suffers from two fatal flaws. 
First, evaluation metric bias. Existing works predominantly frame this as a text generation task evaluated by n-gram matching metrics like ROUGE-L or CIDEr. As demonstrated by G-VEval~\cite{tong2025gveval}, these metrics are highly unsuitable for MLLMs, severely penalizing them for formatting differences rather than genuine perception errors.
Second, limited scope of task types. These datasets strictly focus on visual differences, entirely ignoring the equally critical cognitive dimension of implicit commonalities (\textit{e.g.}, entities that consistently appear across two different scenes). 
Furthermore, their ``minimal changes'' are mostly restricted to salient object attributes within relatively low-resolution contexts, falling entirely short of the extreme ``micro'' scale required for fine-grained perception under complex high-resolution backgrounds.

%% file: sections/3_bench.tex
\section{The Difference Commonality Bench}
\label{sec:bench}

\begin{figure}[tb]
  \centering
  \includegraphics[width=\textwidth]{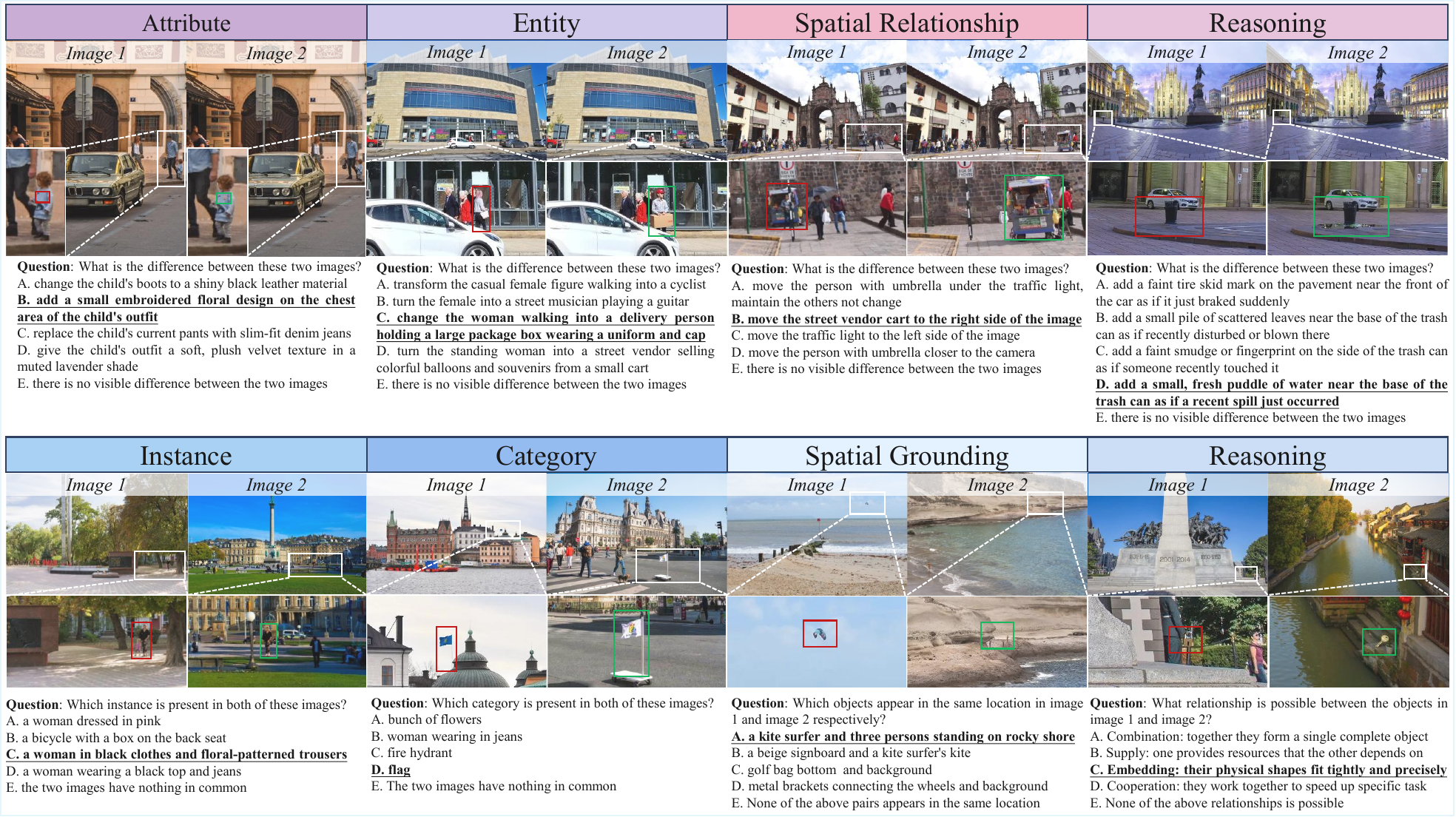}
  \caption{\textbf{Qualitative results on \bench.} The first and second rows illustrate examples of the four task types within the Differential Visual Cues Tasks and Commonality Visual Cues Tasks. Notably, the question for each task contains no explicit text cues. To answer correctly, models must actively perceive the visual cues representing differences or commonalities directly from the image pair.
  }
  \label{fig:samples}
\end{figure}

\bench is designed to address the critical gap in evaluating ``visual-cue guided cross-image fine-grained perception'' by establishing the first corresponding comprehensive benchmark. 
Specifically, \bench systematically evaluates MLLMs across two progressive tracks: \textit{Differential Visual Cues} and \textit{Commonality Visual Cues}. The benchmark comprises 765 meticulously constructed high-resolution samples across 8 distinct task categories, challenging models to discover visual cues from image pairs without explicit textual prompts.

\subsection{Task Definition}

\textbf{Differential Visual Cues} evaluates the model's ability to perceive minute changes within highly aligned or complex high-resolution backgrounds. It includes four progressive categories:
\begin{enumerate}
    \item \textbf{Attribute} evaluates the capacity to discern microscopic alterations in fine-grained visual properties (\textit{e.g.}, spectral reflectance, surface material, or morphological shape) of a specific target, strictly preserving its categorical identity and spatial coordinates. This demands precise semantic decoupling and rigorous analysis of localized visual indicators against complex background interference. 
    \textbf{Example:} In an extremely cluttered electronic workbench, a millimeter-scale DIP switch changes from red to green (or flips from ON to OFF). The implicit question forces the model to focus on the attribute state rather than merely detecting the object.
    
    \item \textbf{Entity} measures the proficiency in detecting categorical substitution, anomalous disappearance, or spontaneous emergence of microscopic entities within dense local regions. Success requires meticulous parsing of localized semantic shifts and robust feature discrimination to identify subtle structural or categorical deviations rather than mere pixel-level noise. 
    \textbf{Example:} In a high-definition street view, a distant ``Speed Limit 60'' sign is replaced by a visually similar ``Speed Limit 80''; or a tiny white circular pill in a medicine box is replaced by a white circular button.
    
    \item \textbf{Spatial Relationship} probes the advanced comprehension of physical-world dynamics by tracking the micro-displacement or topological reorganization of an entity, maintaining strict invariance in its identity and intrinsic attributes. This necessitates robust spatial grounding and the ability to map relative positional shifts within a stable contextual framework, distinguishing genuine spatial semantics from trivial viewpoint variations. 
    \textbf{Example:} A small bunch of keys moves from inside a drawer to hanging on the wall. This strictly differentiates from traditional pixel-level differences by emphasizing explicit spatial semantics.
    
    \item \textbf{Reasoning} assesses the pinnacle of visual discrepancy interpretation, where micro-inconsistencies serve not as static pixel changes, but as ``visual traces'' indicative of latent physical interactions or temporal events. This capability demands second-order cognitive operations, integrating causal inference with precise extraction of visual cues (\textit{e.g.}, footprint deposition) to reconstruct unobserved physical occurrences. 
    \textbf{Example:} Across two high-res beach images, a tiny human footprint appears in the corner of Image B; or a microscopic stress crack emerges on the edge of a perfectly intact glass.
\end{enumerate}

\noindent\textbf{Commonality Visual Cues} introduces a novel challenge: finding the unique intersection of entities, locations, or logic across two high-resolution images with completely different global semantics, lighting, and perspectives. In this scenario, the MLLMs are faced with the fundamental challenge of isolating associative cues between objects while navigating through vast amounts of visual interference and disparity.
\begin{enumerate}
    \item \textbf{Instance} evaluates the capability to achieve physical-instance re-identification under extreme scene and viewpoint variance. 
    \textbf{Example:} Finding a specific Swiss Army knife with unique wear marks hidden in both a highly messy student dormitory (Image A) and an outdoor picnic mat (Image B).
    
    \item \textbf{Category} measures the generalization capability to align micro-targets sharing highly specific taxonomic semantics, despite undergoing domain shifts in scale, chromaticity, material composition, or physical state. 
    \textbf{Example:} A short, red, knotted Ethernet cable dropped on a server room floor versus a long, blue, straight Ethernet cable in a recycling station.
    
    \item \textbf{Spatial Grounding} benchmarks the model's aptitude for identifying spatially equivalent correspondences between disparate objects. Instead of matching appearances, it requires the model to determine whether objects located at identical relative positions in Image A and Image B.
    \textbf{Example:} Identifying spatially corresponding objects across disparate scenes, such as a kite surfer and three persons on a rocky shore versus a beige signboard and a kite, where the model must pinpoint the pair occupying equivalent relative locations.
    
    \item \textbf{Reasoning} represents the apex of Commonality Visual Cues tasks, characterized by the absence of any explicit visual or structural overlap. It requires the model to first develop a fine-grained understanding of nearly every detail within both images. Subsequently, based on this comprehensive perception, the model must deduce whether objects across the two images form a specific functional relationship. Currently, this category encompasses four relationship types:
    \begin{itemize}
        \item \textit{Combination:} The objects together form a single complete entity (\textit{e.g.}, a bottle body and its cap).
        \item \textit{Supply:} One object provides essential energy or resources that the other depends on (\textit{e.g.}, a charger and a smartphone).
        \item \textit{Embedding:} The physical shapes of the objects are designed to fit tightly and precisely into each other (\textit{e.g.}, a memory card and a card slot).
        \item \textit{Cooperation:} The objects must work together to accomplish a specific task (\textit{e.g.}, a hammer and a nail).
    \end{itemize}
    \textbf{Examples:} A solitary, weathered door lock in an old alleyway versus a rusted, vintage key discarded on a park bench; despite the complete lack of visual similarity, the model must recognize the functional ``combination'' relationship between the two. Similarly, a high-voltage power outlet on a modern office wall versus a drained laptop battery in a dark drawer exemplifies a ``supply'' relationship, necessitating deep logical reasoning beyond mere visual perception.
\end{enumerate}

\subsection{Dataset Construction Pipeline}

The \bench dataset is constructed through rigorous human supervision and a multi-stage pipeline that leverages state-of-the-art (SOTA) image editing models and MLLMs for semi-automated data synthesis. To ensure high-quality baseline scenes, we source high-resolution base images from V*~\cite{wu2024vstar}, which is widely recognized in the field of fine-grained perception.

\subsubsection{General Image Editing \& MCQ Formulation}
To establish the foundation of our benchmark, we employ an automated local-editing paradigm. 
First, \textbf{Automated Instruction \& Mask Generation:} Based on grounding annotations, we extract micro-target masks and enlarge them by a $2\times$ context margin to ensure smooth blending. We utilize GPT-5.1 to generate 6 diverse, task-aware modification instructions per mask. To ensure diversity and prevent mode collapse, previously generated instructions are iteratively fed back into the MLLM's context window. 
Second, \textbf{Micro-scale Editing:} The instructions and expanded masks are fed into the FLUX.2 Klein model. This ensures that only the intrinsic properties of the micro-target are altered, while the complex high-resolution background remains perfectly preserved. 
Finally, \textbf{Multiple-Choice Construction:} To eliminate penalties arising from text formatting during evaluation, we structure the benchmark as a Multiple-Choice Question (MCQ). For each valid sample, we randomly select four successful image-instruction pairs as options A, B, C, and D. To prevent the model from exploiting linguistic biases to guess the answer, we ensure that the correct answer is balanced across all options for any given text prompt. Furthermore, to ensure logical completeness and mitigate spurious successes through random guessing, we universally append Option E: \textit{``There is no visible difference between the two images''} or \textit{``There is no visible commonality between the two images''}.

\subsubsection{Task-Specific Construction Protocols}
To fulfill the distinct requirements of the 8 sub-tasks, we design customized generation and filtering workflows, deliberately injecting manual verification and fallback mechanisms at vulnerable nodes.

\noindent \textit{Track 1: Differential Visual Cues.}
\begin{itemize}[leftmargin=*, topsep=2pt, itemsep=2pt]
    \item \textit{Attribute:} The MLLMs generates attributes-altering instructions (e.g., color, material). Following FLUX.2 editing, an automated MLLMs check filters out failed edits, followed by human annotators who verify if the categorical identity and spatial coordinates remain strictly unchanged.
    \item \textit{Entity:} The MLLMs drafts instructions for microscopic object substitution or removal. After generation and editing, human experts inspect the local regions to ensure the substituted entity does not introduce semantic contradictions with the surrounding context.
    \item \textit{Spatial Relationship:} The MLLMs generates instructions to displace the target. After editing and MLLMs-based verification, human annotators conduct a strict review. If the MLLMs-generated spatial instructions violate physical constraints (e.g., objects floating in mid-air) or fall out of logical bounds, human experts directly intervene to supplement and rewrite valid spatial instructions.
    \item \textit{Reasoning:} The MLLMs generates instructions to create ``visual traces'' (e.g., stress cracks, footprints). The generated images are rigorously reviewed by humans to ensure the traces are visually realistic and logically imply a latent physical event.
\end{itemize}

\noindent \textit{Track 2: Commonality Visual Cues.}
\begin{itemize}[leftmargin=*, topsep=2pt, itemsep=2pt]
    \item \textit{Instance:} We crop a specific local region from Image A and utilize FLUX.2 Klein's multi-image editing capability to seamlessly blend the exact entity into a structurally disparate high-res Image B. We prepare 6 candidate blends per source image. After manual inspection, if fewer than 4 valid candidate pairs remain, human annotators take over to supervise the generation and supplement the dataset.
    \item \textit{Category:} Building upon the Instance pipeline, we apply local editing to the entity in Image A to explicitly alter its instance-level attributes (e.g., color, material) while preserving its taxonomic category, ensuring the model matches based on conceptual category rather than identical pixels.
    \item \textit{Spatial Grounding:} Images are divided into $6 \times 6$ grids, and each grid is captioned by MLLMs, followed by human correction. We compute text-embedding similarities using OpenAI's \texttt{text-embedding-3} and select pairs of grids with the \textit{lowest} semantic similarity (ensuring maximum background disparity). Finally, the model is queried to determine whether a target pair of objects occupies the corresponding spatial location across the two images.
    \item \textit{Reasoning:} We pre-define four abstract physical/logical relations: \textit{Combination}, \textit{Supply}, \textit{Embedding}, and \textit{Cooperation}. These relations are translated into pairs of visually recognizable objects. After filtering suitable masks in two entirely different images, the respective objects are synthesized. Finally, human experts verify the quality of the synthesized objects and confirm that the intended answer can be accurately deduced from the images.
\end{itemize}

\subsubsection{Strict Human Verification \& Quality Control}
To ensure the dataset acts as a flawless gold standard, all synthesized pairs undergo a final, exhaustive manual review based on four strict criteria: 
(1) \textit{Textual Fidelity:} The modification must be strictly faithful to the instruction; 
(2) \textit{Visual Naturalness:} The edited image must be free of blurriness, artifacts, or boundary degradation; 
(3) \textit{Mutual Distinguishability:} Modifications across different options (A to D) for the same base image must be mutually exclusive and distinguishable; 
(4) \textit{Scale Constraint:} The altered region must remain rigorously ``micro'' (accounting for $<5\%$ of the total image area). 
Any sample failing to meet all four criteria is either routed back to the human fallback mechanism for manual regeneration or permanently discarded.

\input{tables/main}

%% file: tables/main.tex
\begin{table}[t]
\centering
\caption{Performance comparison of SOTA MLLMs on \bench. The highest and lowest scores for each model type across task types are highlighted in blue and red, respectively. The highest performance achieved by the model in each column is indicated in \textbf{bold}.}
\label{tab:main_results}
\resizebox{\textwidth}{!}{%
\begin{tabular}{lcccccccc|c}
\toprule
& \multicolumn{4}{c}{\textbf{Difference Tasks}} & \multicolumn{4}{c}{\textbf{Commonality Tasks}} & \\
\cmidrule(lr){2-5} \cmidrule(lr){6-9}
\textbf{Model} & \textbf{Attr.} & \textbf{Ent.} & \textbf{Spa.} & \textbf{Rea.} & \textbf{Ins.} & \textbf{Cat.} & \textbf{Spa.} & \textbf{Rea.} & \textbf{Avg} \\
\midrule
Human & 97.6 & 98.9 & 98.0 & 97.3 & 99.1 & 99.1 & 98.4 & 97.4 & 98.3 \\
\midrule
\multicolumn{10}{c}{\textit{Closed-source Models}} \\
\midrule
Gemini-3-Pro & \cellcolor{maxblue} \textbf{49.4} & \cellcolor{maxblue} \textbf{89.5} & \cellcolor{maxblue} \textbf{82.0} & \cellcolor{maxblue} \textbf{28.0} & \cellcolor{maxblue} \textbf{72.7} & \cellcolor{maxblue} \textbf{63.6} & \cellcolor{maxblue} \textbf{58.4} & \cellcolor{maxblue} \textbf{36.4} & \cellcolor{maxblue} \textbf{58.1} \\
Gemini-3-Flash & \cellcolor{minred} 25.0 & 55.6 & 40.0 & \cellcolor{minred} 14.3 & 45.5 & \cellcolor{maxblue} \textbf{63.6} & 41.7 &  \cellcolor{maxblue} \textbf{36.4} & 41.9 \\
GPT-4o & 32.9 & 46.3 & 26.0 & 25.33 & 28.2 & \cellcolor{minred} 24.6 & 26.0 &  20.0 & 29.0 \\
GPT-o4-mini & 38.8 & 53.7 & 54.0 & 22.7 &  71.8 &  60.9 & 48.0 & \cellcolor{minred} 17.4 & 46.3 \\
GPT-4.1-mini & 42.4 & 55.8 & 54.0 & \cellcolor{maxblue} \textbf{28.0} & 70.9 &  60.9 & \cellcolor{minred} 25.6 &  20.0 & 44.1 \\
GPT-4.1 &  28.3 & \cellcolor{minred} 30.5 & \cellcolor{minred} 24.0 & 21.3 & \cellcolor{minred}  22.7 & 26.4 & 27.2 &   19.1 & \cellcolor{minred} 25.0 \\
GPT-5 & 35.3 & 54.7 & 48.0 & 22.7 & 66.4 & 53.6 &  36.0 & 22.6 & 42.6 \\
\midrule
\multicolumn{10}{c}{\textit{Open-source Models}} \\
\midrule
Qwen2.5-VL-7B~\cite{bai2025qwen25vl} & 33.0 & 53.7 & 34.0 & 25.3 & 74.6 & 59.1 & 23.2 & 20.0 & 41.0  \\
Qwen2.5-VL-32B~\cite{bai2025qwen25vl} & 22.4 & 35.8 & 38.0 & 20.0 & 45.5 & 48.2 & 21.6 & 20.0 & 31.4 \\
Gemma3-12B~\cite{gemmateam2025gemma3} & 22.4 & 34.7 & 30.0 & 20.0 & 50.0 & \cellcolor{minred} 39.1 & 29.6 & 20.0 & 31.4 \\
Gemma3-27B~\cite{gemmateam2025gemma3} & \cellcolor{minred} 20.0 & 30.5 & 24.0 & 21.3 & 47.3 & 44.6 & 36.8 & 20.0 & 31.9 \\
InternVL3.5-241B-A28B~\cite{wang2025internvl35} & 22.4 & \cellcolor{minred} 28.4 & 36.0 & \cellcolor{minred} 18.7 & 46.4 & 43.6 & 21.6 & 20.0 & \cellcolor{minred} 29.7 \\
Qwen3-VL-8B~\cite{bai2025qwen3vl} & 29.4 & 51.6 & 40.0 & 21.3 & 71.8 & \cellcolor{maxblue} \textbf{63.6} & \cellcolor{minred} 20.8 & 20.0 & 40.3 \\
Qwen3-VL-30B-A3B~\cite{bai2025qwen3vl} & 31.8 & 46.3 & 36.0 & 22.7 & \cellcolor{minred} 21.8 & 51.8 & \cellcolor{minred} 20.8 & 20.0 & 30.9 \\
Qwen3.5-35B-A3B & 28.3 & 42.1 & \cellcolor{maxblue} \textbf{64.0} & 26.7 & 64.6 & 59.1 & 49.6 & 20.0 & 44.1 \\
DeepEyes-7B~\cite{zheng2025deepeyes} & 31.8 & 55.8 & 36.0 & 26.7 & 75.5 & 60.9 & 29.6 & 20.0 & 42.9 \\
Thyme-7B~\cite{zhang2025thyme} & 23.5 & 42.1 & \cellcolor{minred} 22.0 & 20.0 & 66.4 & 53.6 & 24.8 & 20.0 & 35.6 \\
TreeVGR-7B~\cite{wang2025treebench} & \cellcolor{maxblue} \textbf{35.3} & \cellcolor{maxblue} \textbf{57.9} & 34.0 & \cellcolor{maxblue} \textbf{29.3} & \cellcolor{maxblue} \textbf{76.4} & 62.7 & \cellcolor{maxblue} \textbf{58.4} & 20.0 & \cellcolor{maxblue} \textbf{48.8} \\

\bottomrule
\end{tabular}%
}
\end{table}

%% file: sections/5_experiments.tex
\section{Experiments}

In this section, we first describe the experimental setup and the baselines (\S\ref{sec:exp_setup}). Then, we present a comprehensive evaluation of 18 latest SOTA MLLMs (\S\ref{sec:exp_results}). We demonstrate that while humans can answer the questions with high accuracy, \bench poses significant challenges to existing models. Finally, we provide detailed analyses on multiple experimental settings, examining how high-resolution inputs impact the difficulty of perception tasks. We further reveal the performance patterns of humans during fine-grained perception to inspire future research directions, and perform an error analysis of existing models on \bench (\S\ref{sec:exp_analysis}).

\subsection{Experimental Setup}

\label{sec:exp_setup}
\noindent\textbf{Multimodal LLMs: }
We evaluate \bench on 18 recent MLLMs, including state-of-the-art closed-source models such as Gemini-3 (Flash, Pro), and the GPT family (4o, o4-mini, 4.1-mini, 4.1, 5); state-of-the-art open-source models such as Qwen3-VL (8B, 30B-A28B)~\cite{bai2025qwen3vl}, Qwen2.5-VL (7B, 32B)~\cite{bai2025qwen25vl}, Gemma3 (12B, 27B)~\cite{gemmateam2025gemma3}, InternVL3.5 (241B-A28B)~\cite{wang2025internvl35}, and Qwen3.5; as well as models specifically optimized for fine-grained perception, including DeepEyes~\cite{zheng2025deepeyes}, Thyme~\cite{zhang2025thyme}, TreeVGR~\cite{wang2025treebench}, and Mini-O3~\cite{lai2025minio3}.


\vspace{1ex}\noindent\textbf{Evaluation setup: }
We follow standard setups as in the VLMEvalKit~\cite{duan2024vlmevalkit}, where the temperature is set to 0. Specifically, we require the models to directly output the corresponding answer letters for the MCQs (A, B, C, D and E) and evaluate them using exact letter matching. We find that in most cases, existing models demonstrate strong instruction-following capabilities.

\vspace{-3mm}
\subsection{Main Results}
\label{sec:exp_results}
\noindent\textbf{Overall Performance: }
\bench poses a formidable challenge to existing MLLMs. As shown in \Cref{tab:main_results}, even the most advanced closed-source models (e.g., Gemini-3-Pro) and open-source models (e.g., TreeVGR-7B) achieve average accuracies of only 58.1\% and 48.8\%, respectively, indicating a massive performance gap compared to the 98.3\% average accuracy of humans. In particular, most models perform poorly in the reasoning (Rea.) sub-tasks within the Difference Tasks category, with scores for many models falling below 20-30. Notably, the reasoning tasks within Commonality Tasks present an exceptionally high level of difficulty. Although these tasks are easily solvable for human participants, we observe that nearly all existing open-source models mistakenly perceive no commonalities between the two images, thus predominantly selecting Option E, which leads to trivialized, identical results. This highlights a significant limitation in existing models when handling fine-grained cross-image comparison.

\noindent\textbf{Closed-source vs. Open-source Models: }
Within the closed-source camp, Gemini-3-Pro leads with an average score of 58.1\%, demonstrating strong capabilities in entity (Ent.) and spatial (Spa.) understanding for Difference Tasks. In contrast, the performance of the GPT series is highly inconsistent; for instance, GPT-4.1-mini excels in attribute (Attr.) recognition within Difference Tasks (42.4\%) but scores only 25.6\% in spatial (Spa.) understanding within Commonality Tasks. This reveals an uneven distribution of capabilities across different visual reasoning tasks among closed-source models. Regarding open-source models, TreeVGR-7B demonstrates remarkable competitiveness, with its 48.8\% average score outperforming several closed-source models (e.g., GPT-4o at 29.0\% and GPT-4.1 at 25.0\%). This underscores the potential for lightweight, task-specific optimized models to compete effectively with large-scale closed-source counterparts.

\begin{wrapfigure}{l}{0.5\textwidth} 
    \centering
    \includegraphics[width=0.48\textwidth]{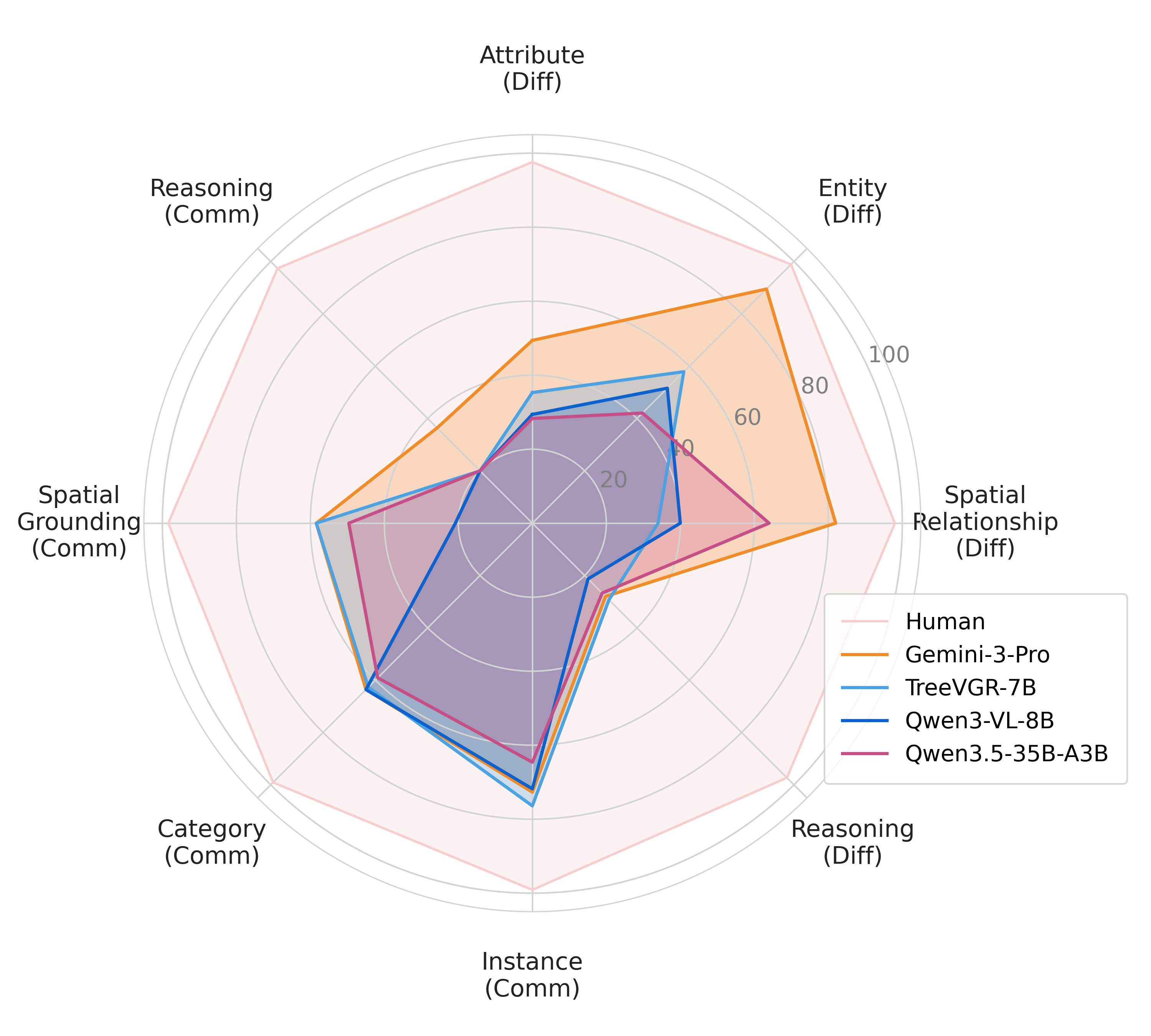}
    \caption{Accuracies of MLLMs on \bench.}
    \label{fig:radar}
    \vspace{-25pt}
\end{wrapfigure}


\noindent\textbf{Performance Disparities across Tasks: }
A comparison across task types reveals that the categorization (Cat.) task within Commonality Tasks is a relative strength for current models (with most scoring above 50\%), whereas the reasoning (Rea.) task remains a universal ``Achilles' heel,'' with most models hovering around 20\% in \Cref{fig:radar}. This phenomenon suggests that while current MLLMs have made progress in fundamental object recognition and classification, they still lack the ability to capture complex visual cues required for deep logical analysis and multi-image fine-grained perception.

\begin{figure}[t]
    \centering
    \begin{minipage}[t]{0.44\textwidth}
        \centering
        \includegraphics[width=\textwidth]{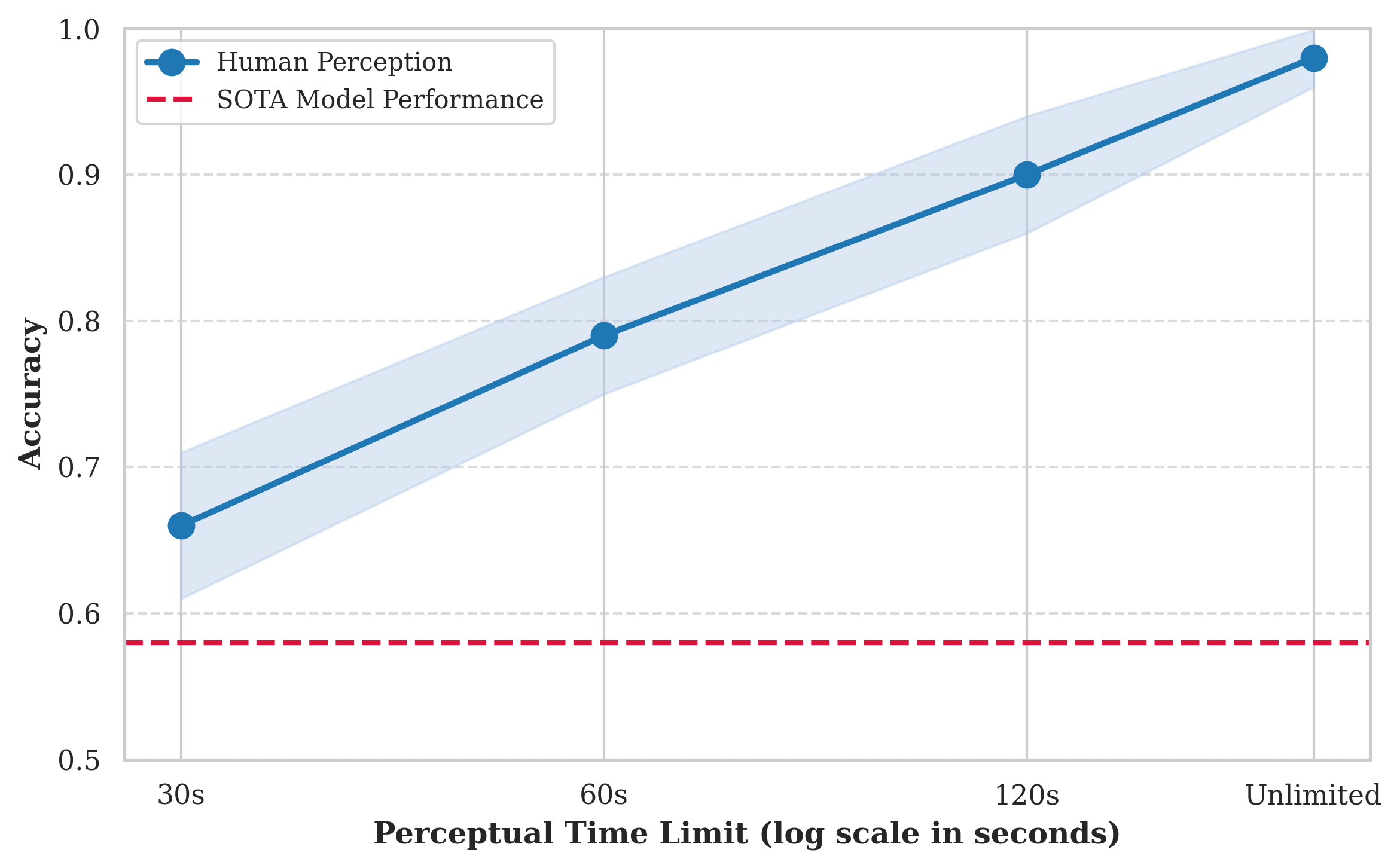}
        \caption{Human accuracy on \bench as a function of perceptual duration. The SOTA model Gemini-3-Pro is shown for comparison.}
        \label{fig:human_acc}
    \end{minipage}
    \hfill 
    \begin{minipage}[t]{0.55\textwidth}
        \centering
        \includegraphics[width=\textwidth]{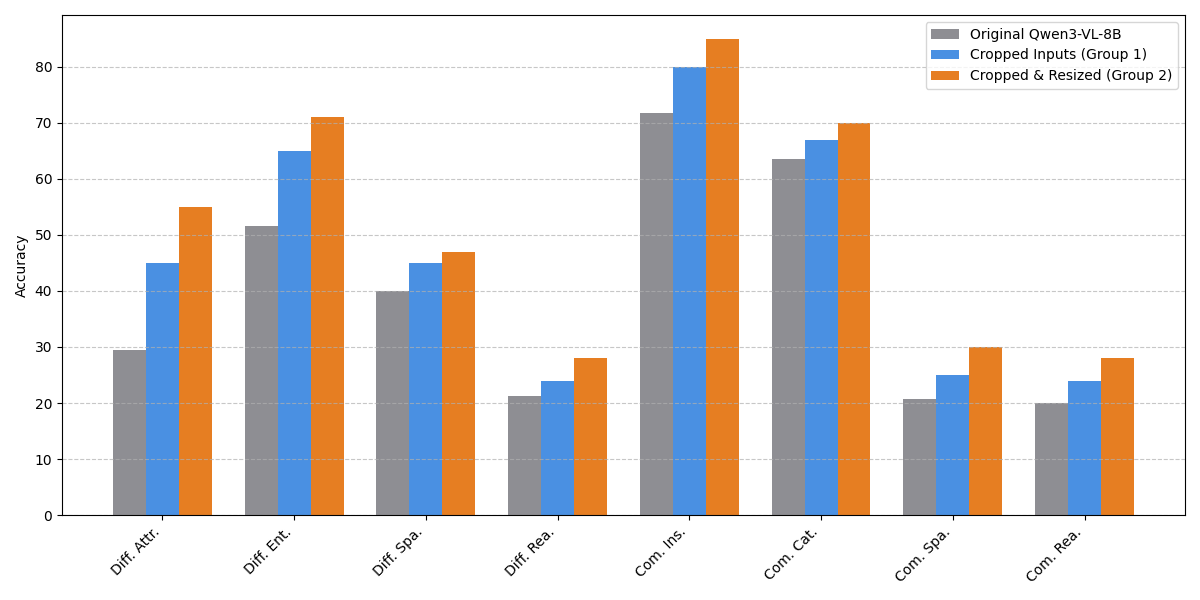}
        \caption{Performance comparison of the Qwen-3-VL model on \bench under different settings: vanilla inputs, cropped inputs, and resized inputs.}
        \label{fig:compare_acc}
    \end{minipage}
\end{figure}

\noindent\textbf{Is human perception instantaneous?}
Although existing benchmarks commonly include human performance result, the relationship between human perceptual processes and accuracy has been absent, particularly in fine-grained perception tasks. We attempt to analyze and demonstrate this relationship on \bench by tracking human perceptual duration against accuracy. We invited eight Ph.D. students, who had no prior exposure to the test, to serve as participants. Each two were required to complete the tasks under one of the four time constraints: 30s, 60s, 120s, and unlimited time; the final results represent the average. As shown in \Cref{fig:human_acc}, we find that human precision in perception is not instantaneous; when perceptual time is limited, the performance gap between humans and state-of-the-art models begins to narrow. However, as the human time investment increases, perceptual accuracy improves consistently, eventually reaching a near-perfect state. This suggests that the perceptual capabilities of existing MLLMs, like their reasoning abilities, may benefit from increased computational investment.

\subsection{Analysis}
\label{sec:exp_analysis}

\noindent\textbf{How does high resolution challenge fine-grained perception?}
Compared to low-resolution perception tasks, high-resolution images, on one hand, increase the volume of visual information that models must process; on the other hand, they lower the relative scale at which visual cues must be identified to be effectively observed. To explore how high resolution challenges existing MLLMs, we conducted comparative experiments using the Qwen-3-VL-8B model. We sampled $1/10$ of the instances from each task in \bench, annotated the positions of the visual cues manually, and performed two sets of controlled experiments. In the first set, we used the cropped images based on the annotated positions as inputs. In the second set, we cropped the original images and then resized them back to the original image dimensions as inputs.
As shown in \Cref{fig:compare_acc}, the first set demonstrates a significant improvement compared to the original evaluation results. Notably, the second set achieves an even greater performance gain compared to the first. These results indicate that, on one hand, filtering out irrelevant visual regions significantly enhances model performance, suggesting that high-resolution images increase evaluation difficulty by introducing an excessive volume of extraneous visual information. On the other hand, high resolution poses the challenge of excessively low visual cue ratios. Therefore, increasing the relative proportion of these cues contributes substantially to improved perceptual performance.

\begin{wrapfigure}{R}{0.5\textwidth} 
    \centering
    \includegraphics[width=0.48\textwidth]{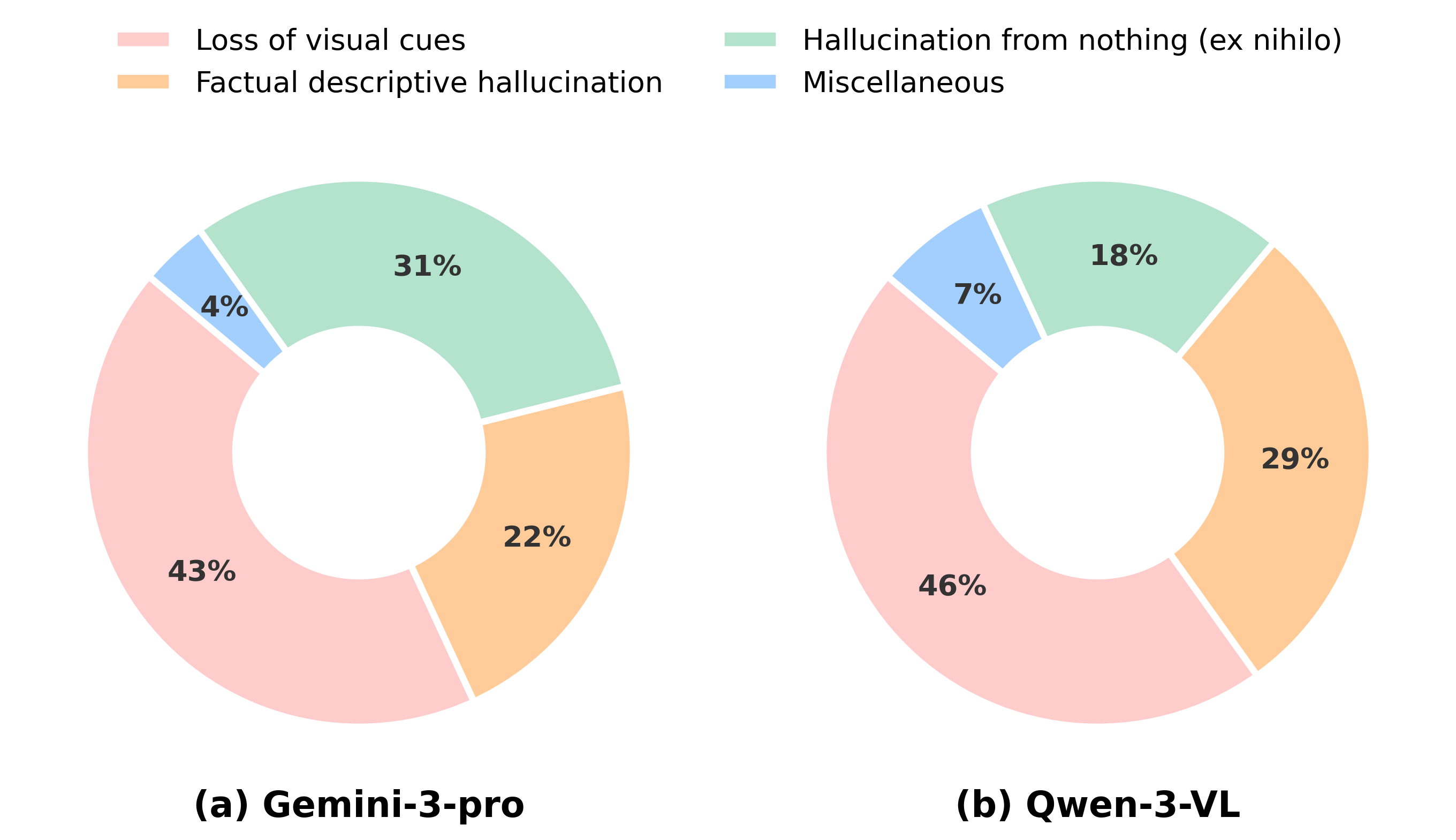}
    \caption{Proportional distribution of error types for Gemini-3-Pro and Qwen-3-VL on \bench.}
    \label{fig:error}
\end{wrapfigure}

\noindent\textbf{Error analysis:}
To investigate the reasons behind the potential failures of existing models, we employed Gemini-3-pro and Qwen-3-VL as representatives of closed-source and open-source models respectively. Since our evaluation protocol requires models to output the final answer directly, which hinders the assessment of their reasoning processes, we allowed the models to generate brief descriptions before providing the answers during the error analysis. We categorized the errors into four types:
\begin{enumerate}
\item \textit{Loss of visual cues:} Ignoring visual cues regarding the differences or commonalities between two images.
\item \textit{Factual descriptive hallucination:} Misjudgments such as the orientation of a dog or the position of a person in the images.
\item \textit{Hallucination from nothing (ex nihilo):} Perceiving differences between identical images, or identifying commonalities where none exist.
\item \textit{Miscellaneous:} Refusals to answer or errors with indeterminate causes.
\end{enumerate}
As shown in \Cref{fig:error}, we found that although there is a significant performance gap between Gemini-3-pro and Qwen-3-VL, their error distributions are quite similar. The loss of visual cues is the primary issue for existing models, suggesting a fundamental problem in these models regarding visual-cue-guided perception tasks that have not been evaluated in the past, likely due to a systemic lack of training data. Beyond the loss of visual cues, we observed distinct hallucination preferences: Gemini-3-pro is more prone to ``hallucination from nothing,'' while Qwen-3-VL tends to produce ``factual descriptive hallucinations.'' These errors indicate that future MLLMs should prioritize optimization for recognizing visual cues in perception tasks to prevent the neglect of subtle visual information. Concurrently, it is also essential to enhance general perceptual accuracy to mitigate both factual descriptive hallucinations and hallucinations from nothing.

%% file: sections/6_conclusion.tex
\section{Conclusion}
In this paper, we have presented \bench, the first comprehensive benchmark tailored to evaluate high-resolution, cross-image fine-grained perception in MLLMs. By synthesizing datasets that necessitate the detection of implicit visual cues without explicit textual guidance, we have demonstrated that current SOTA models, both closed-source and open-source still struggle particularly in tasks requiring  the integration of micro-scale visual information. Our analysis highlights that while these models have achieved proficiency in fundamental object recognition, they still struggle significantly with the proactive interpretation of spontaneous visual discrepancies and commonalities. We believe that \bench not only provides a rigorous framework to expose these limitations but also offers a critical frontier for developing more robust, perceptually aware multimodal systems capable of bridging the gap between current machine performance and human-level visual cognition.